\documentclass{article}

\usepackage{times}
\usepackage{graphicx} % more modern
\usepackage{subfigure} 

\usepackage{natbib}

\usepackage{algorithm}
\usepackage{algorithmic}

\usepackage{hyperref}

\usepackage[accepted]{icml2016}

\icmltitlerunning{Deep Learning-Based Image Kernel for Inductive Transfer}

\begin{document} 

\twocolumn[
\icmltitle{Deep Learning-Based Image Kernel for Inductive Transfer}

\icmlauthor{Neeraj Kumar, Animesh Karmakar, Ranti Dev Sharma, Abhinav Mittal, Amit Sethi \\}{(neeraj.kumar,a.karmakar,abhinav.mittal,ranti,amitsethi)@iitg.ernet.in}
\icmladdress{Indian Institute of Technology Guwahati,
            Assam, India, 780139}
%\icmlauthor{Your CoAuthor's Name}{email@coauthordomain.edu}
%\icmladdress{Their Fantastic Institute,
   %         27182 Exp St., Toronto, ON M6H 2T1 CANADA}

\icmlkeywords{boring formatting information, machine learning, ICML}

\vskip 0.3in
]

\begin{abstract} 

We propose a method to classify images from target classes with a small number of training examples based on transfer learning from non-target classes. Without using any more information than class labels for samples from non-target classes, we train a Siamese net to estimate the probability of two images to belong to the same class. With some post-processing, output of the Siamese net can be used to form a gram matrix of a Mercer kernel. Coupled with a support vector machine (SVM), such a kernel gave reasonable classification accuracy on target classes without any fine-tuning. When the Siamese net was only partially fine-tuned using a small number of samples from the target classes, the resulting classifier outperformed the state-of-the-art and other alternatives. We share class separation capabilities and insights into the learning process of such a kernel on MNIST, Dogs vs. Cats, and CIFAR-10 datasets.

\end{abstract}

\section{Introduction}

Deep learning architectures, notably the variants of convolutional neural networks (CNNs), have produced state-of-the-art results in large as well as very large-scale image classification and recognition problems ~\cite{Hinton1, Simonyan14}. The success of CNNs can be attributed to their capabilities of learning a hierarchy of increasingly complex and class-specific features using efficient training algorithms with appropriate connectivity and weight-sharing constraints in the convolutional layers. However, the practical utility of CNNs is limited in situations where training data is limited due to their large training sample requirement to achieve acceptable recognition rates. For instance, some of the top performing variant of CNN on CIFAR-10 dataset use 50,000 training samples and take a few days to train on a single workstation~\cite{Hinton1}. In this paper, we focus on reducing the training sample and time requirement of CNN-based techniques while still hoping to utilize their hierarchical feature learning capabilities for high classification accuracy.

Before the explosion of deep learning, use of support vector machines (SVMs) with various hand-crafted features represented the state-of-the-art for image recognition~\cite{vapnik}. Although, SVM-based systems produced lower peak recognition rates, their main advantage was lower number of training sample and time requirement than their deep learning counterparts~\cite{Hinton1}. SVMs have been coupled with variants of CNN by replacing the latter's classification layer (usually a softmax layer). In some cases, SVM not only functions as a wide margin classifier, it also provides cost and gradient values for CNN training. But due to gradient dilution at the lower layers of CNN, a large number of training samples and training time are still required for good results~\cite{Hinton1}. It has also been shown that CNN features\footnote{CNN features is a term that usually means a flattened vector of the output of the final convolutional layer of a CNN.} that was pre-trained in a supervised manner on non-target classes\footnote{We assume that we are interested in classifying samples from a \emph{target} set of classes with a small number of labeled samples, while we have access to a large number of labeled examples from separate set of \emph{non-target} classes for transfer learning.} can be used as an image representation in an SVM for target classes to give surprisingly good recognition performance~\cite{huang06}. Such transfer learning from non-target classes is our main inspiration for experimenting with a different architecture that improves upon their recognition rates for a small number of training samples from target classes.

We propose an SVM-based classifier that operates on a trainable kernel based on a Siamese deep neural network (henceforth, Siamese net)\footnote{A Siamese network usually has an identical pair of convolutional and pooling layer stacks that share a single stack of fully connected layers on top. It is mainly trained and used to compute similarity between the pair of input images.}~\cite{Sumit}. Assuming lack of a notion of semantic similarity, we propose that the Siamese net can be trained to estimate the probability that its input image pair belongs to the same class, \emph{no matter what that class is}. Our hope is to be able to apply such Siamese net to target classes with no to little fine-tuning, which requires that its learning of similarity using non-target classes generalize to target classes.

Although many transfer learning methods for CNNs also use only the class labels for pre-training on non-target classes, the hope in CNNs is to learn discriminative features that generalize from non-target classes to target classes. We compare these two learning approaches in terms of their image recognition performance on standard datasets -- MNIST digits~\cite{mnist}, Dogs vs. Cats~\cite{catdog}, and CIFAR-10 objects~\cite{cifar}. We also experimented with two different levels of transfer learning -- one in which the fully connected layers on top of the Siamese convolutional stacks are fine-tuned on target classes, and another in which even the fully connected layers are trained on non-target classes. In both cases, we reduced the practical training time by not training the convolutional layer at all, and copying weights from CNNs trained by others. Yet, in most cases, our kernels outperformed static kernels (RBF and linear) operating on CNN features.

For use in an SVM, the output of the Siamese net is not guaranteed to be positive semi-definite (PSD), which is one of the Mercer’s conditions~\cite{scholkopfkernel}. To remedy this, we propose to use a transformation of the gram matrix produced by the Siamese net to a PSD matrix. We refer to this kernel as DEep Semantic Kernel (DESK).

\section{Related Work}

CNNs, starting from LeNet~\cite{lenet}, ushered the popularity of deep learning by improving object recognition rates significantly over previous approaches. Their main benefit is that they do not rely on hand-crafted features and learn an increasingly complex feature hierarchy from the data itself. To improve the image recognition performance of CNNs, various changes in hyperparameters and the mathematical functions in each layer and their training algorithms have been proposed~\cite{pooling,relu,Dropout}. It has also been demonstrated that a linear SVM used in the final layer instead of logistic sigmoid for binary classification may improve the recognition performance~\cite{largescale}. However, the number of training samples required for even these variants of CNN remains large because they retain the need to train a deep stack of layers using gradient descent, where the gradient dilutes away from the output layer~\cite{Hinton1}.

Use of unlabeled or labeled data from non-target classes in a transfer learning setting has become a common practice in CNNs~\cite{huang06,bachman14,alexey15}. Practical ability to utilize transfer learning has increased with the advent of deep learning libraries such as Theano~\cite{theano}, Caffe~\cite{caffe}, and researchers' willingness to make their models trained using these libraries accessible online. Transfer learning is done mainly by simply copying or initializing the weights of initial layers that are most affected by gradient dilution using models trained on other classes. It has been shown that the first few layers learn class-agnostic features that are most amenable to transfer learning~\cite{bengio}. An SVM trained on CNN features obtained using non-target classes gives decent classification performance out of the box~\cite{bengio}. Other approaches to transfer learning in CNNs besides using pre-trained convolutional layers include the use of successive frames of a video for unsupervised pre-training~\cite{AngVideo}. However, for state-of-the-art performance layers are still fine-tuned based on tens, if not hundreds, of thousands of training examples from target classes, which also reflects in their training times.

With DESK, we take the approach of learning a kernel instead of features for use in SVM using Siamese nets -- a variant of CNN. A Siamese net, such as the one used in DESK, computes similarity $s_{ij}$ between images $i$ and $j$. This approach has been used in recent works to learn distance between faces for verification~\cite{hu2014} and similarity learning between image patches for wide baseline stereo matching~\cite{Sergey}. With DESK, we went beyond similarity learning to propose a complete kernel-based classification framework that also utilizes transfer learning to yield high classification accuracy with a small training sample size.

Kernel learning outside of deep learning has been an active area of research. Most notable successes have been multiple kernel learning~\cite{Soren},~\cite{Mehmet}, hierarchical arc-cosine kernels~\cite{Youngmin1}, and use of semi-definite programming to learn a kernel matrix~\cite{semidefinite}. While these works have shown improvement over the use of static kernels for image recognition, due to their reliance on hand-crafted and shallow features, these have been outperformed by their deep learning counterparts with the exception of~\cite{Youngmin1}.

Attempt to show equivalence of hierarchical kernels and deep learning include~\cite{Muller} and~\cite{Youngmin1}, while use of multiple kernels in a convolutional architecture was proposed by~\cite{Cordelia}. These works further the theoretical understanding of deep learning. Our attempt is in line with these efforts but is aimed at reducing the training time and sample requirement of CNNs without compromising on object recognition accuracy.

\section{DESK Architecture and Training}

With DESK, our goals in addition to learning similarity between images from target classes were the following:

\begin{enumerate}
\item \emph{Propose an efficient method} to learn to compute similarity between pairs of images from target classes in terms of sample and time requirement.
\item \emph{Study trade-off} between the extent of transfer learning from a kernel trained on non-target classes and target classes and classification accuracy.
\item \emph{Probe for differences} between learning a deep kernel and deep features for classification.
\item \emph{Convert the similarity score into a Mercer kernel} to train an SVM for classification of target classes.
\end{enumerate}

\begin{figure}[!htb]
\centering
\includegraphics[width=0.5\textwidth, keepaspectratio]{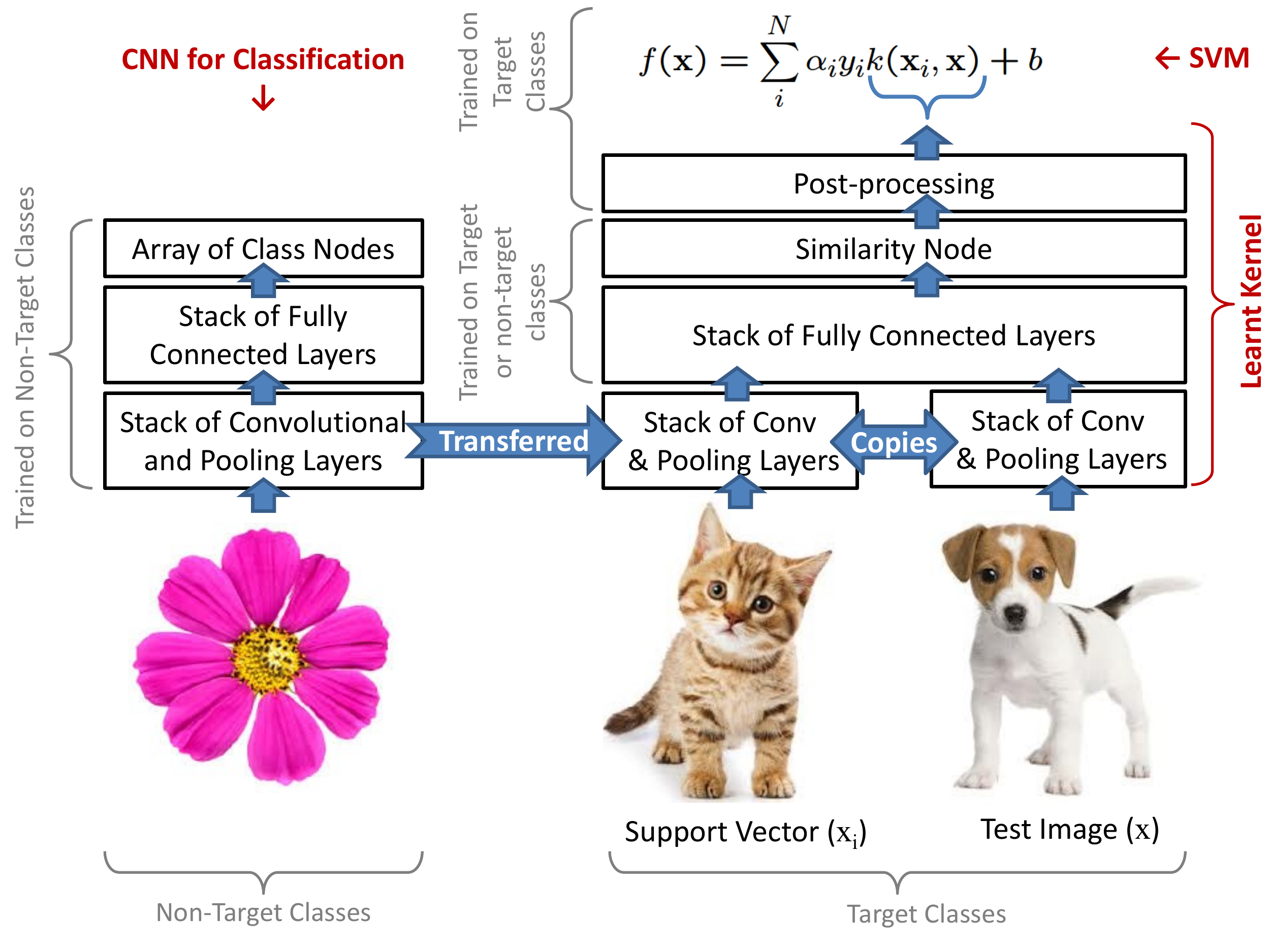}
\caption{Proposed classification scheme based on an SVM using a trainable kernel that lends itself to transfer learning.}
\label{Fig:Abstract}
\end{figure}

DESK's Siamese net architecture takes two images to be compared as inputs and computes their similarity. This similarity score is converted into a Mercer kernel using a post-processing step. The kernel is then used to train an SVM using input images from target classes. This is shown in Figure \ref{Fig:Abstract}. We start with presenting different options for DESK's neural network (NN).

\subsection{NN architecture}
The defining features of the architectures that we experimented with were the following:

\begin{enumerate}

\item \emph{Input}: Two images, $i$ and $j$, to be compared for similarity were taken as input.
\item \emph{Output}: A target output for supervised training that was a scalar (similarity score), unlike a vector in case of multi-class CNNs.
\item \emph{Convolutional layers}: A stack of convolutional layers (includes convolution, nonlinear squashing, and pooling) extracted increasingly complex features from the input images.
\item \emph{Fully connected layers}: A stack of fully connected layers was used between the stack of convolution layers and the scalar output node trained to contribute to similarity estimation.
\end{enumerate}

\begin{figure}
\centering
\includegraphics[width=0.47\textwidth]{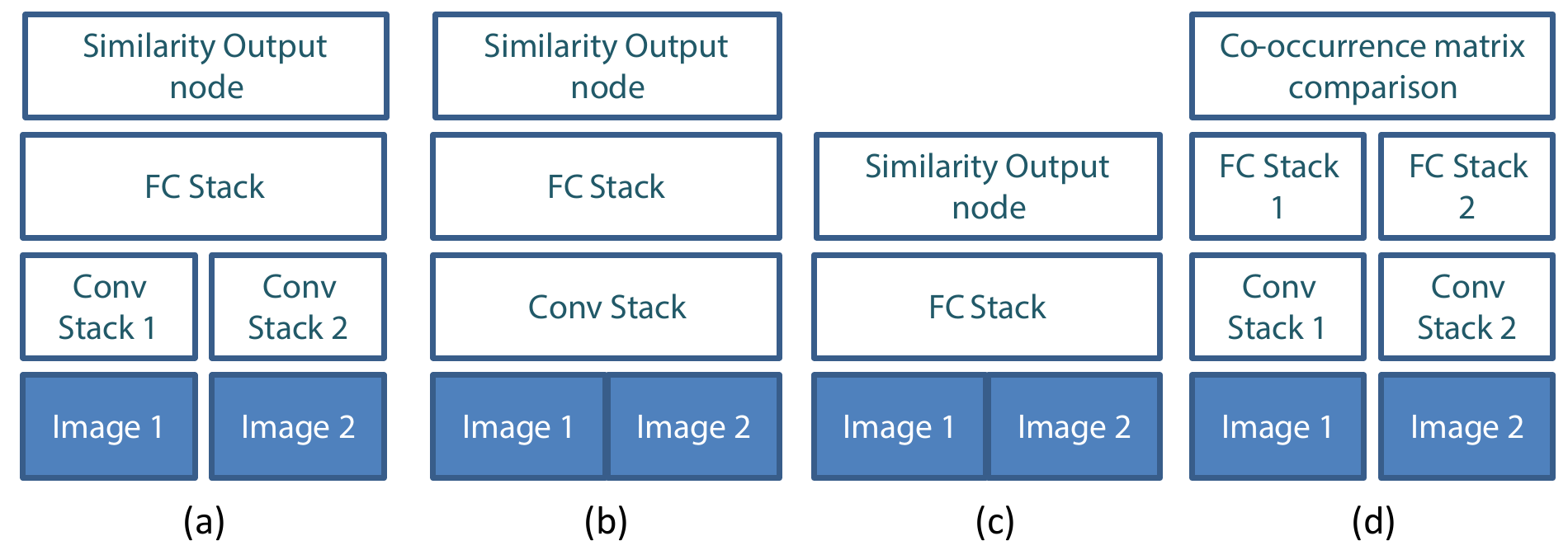}
\caption{Various possible architectures for the neural network of the trainable kernel: (a) Siamese, (b) Super-Image, (c) Only fully-connected layer, (d) Deep-feature co-occurrence matrix.}
\label{Fig:Archs}
\end{figure}

For DESK, the most successful architectures were \emph{Siamese}~\cite{Sumit}, where the two input images were processed using two paired stacks of convolutional layers that had disjoint connections but shared (identical) weights, followed by a common stack of fully connected layers. Other architectures that we tried but found to be not as well-suited for estimating image pair similarity are shown in Figure \ref{Fig:Archs}. For other architectures that might work, the reader is referred to~\cite{Sergey}.

\subsection{Levels of transfer learning}
\label{sec:trainingDesc}

To address our first goal of practical efficiency in training a kernel, we decided to not train the convolutional layers of the Siamese net from scratch at all. We simply picked pre-trained CNNs trained to classify an appropriate set of non-target classes and duplicated their convolutional layers in the two paired stacks of convolutional layers in the Siamese net. We did not fine-tune these layers at all after that. This level of transfer learning was common to all of our experiments. We experimented with two  transfer learning schemes or levels:

\begin{enumerate}

\item \textbf{Conv-transfer:} In the first scheme, after copying the convolutional layers from non-target classes and freezing them, we trained the shared stack of fully connected layers on pairs of images (processed through the convolutional layers) from \emph{target} classes. This was done by either fine-tuning fully connected layers pre-trained on image pairs from non-target classes by using a image pairs from target classes, or training from scratch on target classes.

\item \textbf{Full-transfer:} In the second scheme, in addition to copying convolutional layers from non-target classes and freezing them, we also trained the shared stack of fully connected layers on pairs of images (processed through the paired conv layers) from \emph{non-target} classes. Then we froze the entire Siamese net while processing the target classes.

\end{enumerate}

 Interestingly, even the conv-transfer learning scheme requires only a small number of samples from target classes to train the fully connected layers because a small number of images can generate a large number of image \emph{pairs}.

\subsection{Supervised training of Siamese net}

The notion of similarity that we propose is the probability of an image pair to belong to the same class no matter what their classes are. The main reason for choosing this notion of similarity as opposed to, say, semantic distance between class labels words, was to facilitate comparison with CNNs that use the same supervised information. We wanted to test the hypothesis that \emph{learning to compare is inherently a better objective than learning to classify in terms of training sample and time efficiency.}
 
We trained the shared fully connected layers of the Siamese nets with CNN features for a pair of images computed using frozen conv layers as inputs. Supervised target output of $\{0,1\}$ was emulated during training, where 1 was used to code similar image pairs (belonging to the same class), and 0 for dissimilar pairs (belonging to different classes).  We used an equal number of similar and dissimilar pairs for training.

\subsection{From Siamese net to kernel}

According to Mercer's theorem, a function of two inputs represents a reproducing kernel Hilbert space if it is (a) symmetric, and (b) positive semi-definite (PSD)~\cite{scholkopfkernel}. The Siamese net output in DESK is not guaranteed to be that function, which can occasionally cause convergence problems for SVM packages training on gram matrices that are not PSD. Therefore, before training an SVM, which requires such a kernel for \emph{guaranteed} convex objective function, we experimented with the following post-processing schemes for the gram matrix $S$ with elements $s_{ij}$:

\begin{enumerate}

\item{\bf Co-incidence matrix:} It has been suggested that when a non-negative similarity metric captured in gram matrix $S$ is asymmetric, it can be converted to a co-incidence matrix $S^{T}S$ that is PSD and can also be used for classification in an SVM~\cite{representer}. The added advantage of this scheme in processing unseen target classes is that computing co-incidence confers another level of learning where each sample's vector of distances with other samples is compared to that of other vectors. So, it transforms $S$ into a matrix $S^{T}S$ that really is tailored to the inter-distances from the samples that generated of $S$ itself. In our experiments, this scheme gave consistently high separation of test cases and high classification performance.

\item{\bf Pair exchange and ignoring negative eigen-values:} First we exchanged the inputs and took the average of the two kernel computations, that is, we used the matrix $\frac{1}{2}(S+S^{T})$. Then we computed eigen decomposition of this symmetric matrix, set the negative eigenvalues to zero, and back-projected the eigenvectors using the new eigenvalues to obtain a symmetric and PSD matrix. We used this matrix for training the SVM, and used the projection of the symmetrized testing gram matrix into this eigenspace for testing the SVM. This scheme didn't perform as well as the co-incidence scheme.

\item {\bf Pick-out kernel:} Munoz et. \emph{al.} presented a method, known as \emph{pick-out}, that specifically takes classification labels into account to build the proximity matrix for enforcing positive semidefiniteness on the asymmetric Gram matrices~\cite{pickout}.  In this method, while training, the $max(s_{ij},s_{ji})$ is used as the kernel value if both samples (images) $i$ and $j$ belong to the same class, and $min(s_{ij},s_{ji})$ otherwise. For testing, one first assumes that the test sample belongs to one class to compute its distance from the separating hyperplane, and then repeats the exercise with assuming that the test belongs to the other class. The binary class decision is taken based on comparison of the two signed distances. Its extension to more than two classes isn't obvious, and it also gave worse performance than co-incidence matrix for Dogs vs. Cats binary classification.

\end{enumerate}

\subsection{SVM training}

For obvious reasons, the SVM was always trained using kernel matrices from target classes. We tested its performance on kernels derived from conv-transfer and full-transfer schemes for different number of training samples. For conv-transfer, we used the same samples to train the fully-connected layers of the Siamese net and the SVM. SVM training took around 5 to 10 minutes.

\section{Experiments and Results}

In this Section, we describe the data used for target classes, and their non-target classes as well as the results of our experiments.

\subsection{Data sets and DESK architectures}

We probed questions around the properties of the learned kernel as well as the recognition performance when the kernel was used in SVM. To do so, we selected three datasets of target classes along with their respective non-target classes. The non-target classes were selected such that the images were of the same type as those of the target classes in terms of size and broad categories. For example, two hand-written scripts share pen strokes, while two sets of natural images share similar features and feature hierarchy. This pairing of target and non-target classes is shown in Table \ref{tab:TargetNonTarget}.

\begin{table}[!h]
\caption{Target and non-target classes}
\label{tab:TargetNonTarget}
\centering
\begin{tabular}{|p{3cm}|p{0.4cm}|p{3cm}|}
\hline
Target classes & \# & Non-target classes \\ \hline
Dogs vs. Cats ~\cite{catdog} & 2 & ImageNet sans dogs and cats ~\cite{imagenet} \\ \hline
Handwritten digits (MNIST) ~\cite{mnist} & 10 & Handwritten alphabet (NIST) ~\cite{nist} \\ \hline
CIFAR-10 objects ~\cite{cifar} & 10 & CIFAR-100 objects ~\cite{cifar} \\ \hline

\end{tabular}

\end{table} 

We selected cat vs. dog classification task because it is binary and gives better insights into the performance of the kernel and the SVM without having to interpret how these scale to multi-class problems. It should be noted that this is not a trivial task because both dogs and cats are furry mammals usually pictured in similar surroundings. While the other two target datasets and their non-target counterparts had a fixed image size, ImageNet had variable image sizes. We standardized the images in ImageNet and to 227$\times$227$\times$3 by scaling.

The best performing architectures for the two of the datasets used in our experiments as defined by kernel accuracy are shown in Table \ref{tab:Architectures}, where FC represents fully connected layers whose number of neurons are mentioned, and Conv represents Siamese convolutional and pooling layers whose filter sizes, number of filters, and pooling sizes respectively are mentioned. Note that all nonlinearities were rectified linear units (ReLU), except for the output node, which had a logistic sigmoid nonlinearity. We used AlexNet pre-trained architecture for ImageNet~\cite{alexnet}. For CIFAR dataset, we used mxnet architecture~\cite{mxnet}, but we do not show it here because it is highly complex. The convolutional layers for NIST were trained from scratch using a CNN that gave approximation 92\% classification accuracy on English alphabet. The FC layers were trained using hinge loss cost function as suggested in~\cite{Sergey} and used a dropout of 0.5.

\begin{table}[!h]
\caption{Neural network architectures. In the convolutional layers, filter size, number of filters, and pooling size are mentioned respectively.}
\label{tab:Architectures}
\centering
\begin{tabular}{| c | c | c |}
\hline
Data set & ImageNet & NIST \\ \hline
FC 3& 1,000& \\ 
FC 2& 4,096& \\ 
FC 1& 4,096& 800 \\ \hline
Conv 5& 3$\times$3, 256, 3$\times$3& \\ 
Conv 4& 3$\times$3, 384, 1$\times$1& \\ 
Conv 3& 3$\times$3, 384, 1$\times$1& 3$\times$3, 128, 2$\times$2 \\ 
Conv 2& 5$\times$5, 256, 3$\times$3& 3$\times$3, 64, 2$\times$2 \\ 
Conv 1& 11$\times$11, 96, 3$\times$3& 3$\times$3, 32, 2$\times$2 \\ \hline
Image & 227$\times$227$\times$3& 32$\times$32$\times$1 \\ \hline

\end{tabular}

\end{table} 

\subsection{Kernel performance}

We tested the kernel's performance for both transfer schemes described in Section~\ref{sec:trainingDesc}. While conv-transfer was our main focus, full-transfer shows the generalization capabilities of features estimated for learning to compare as opposed to learning to classify.

\subsubsection{Conv-transfer}
Our main experiments were about only transferring the convolutional layers. Then, we trained the fully connected layers on training images (actually, pairs thereof) from the target classes using 500, 1,000 and 5,000 images. This represents the conv-transfer learning scheme. However, as shown in Table~\ref{tab:KernelValidation}, we used a lot more pairs of images that we could generate from the paltry number of images. We then tested the kernel on 10,000 pairs from a held-out set of 1,000 images from the target classes. We generated an equal number of similar and dissimilar pairs for both training and testing.

The kernel accuracy decreased as the number of training images was increased. However, this lower kernel performance due to more extensive training need not necessarily lead to worse recognition performance, as the kernel was trained to find a more meaningful notion of similarity on a larger set of images. Kernel accuracy can give a crude approximation of 1-shot learning performance for a binary classification problem.

\begin{table}[!h]
\caption{Kernel testing AUC for training fully connected layers on the different number of samples from target classes.}
\label{tab:KernelValidation}
\centering
\begin{tabular}{| r | r | r | r | r |}
\hline
Images& Pairs& Dogs Cats& MNIST& CIFAR-10 \\ \hline
500& 10,000& 0.999& 0.996& 0.996 \\ 
1,000& 60,000& 0.999& 0.994& 0.999 \\ 
5,000& 60,000& 0.999& 0.993& 0.997 \\ \hline

\end{tabular}

\end{table}

\subsubsection{Full-transfer}

To test generalization capabilities of the kernel, we tested the first transfer learning scheme described in Section~\ref{sec:trainingDesc} by training the fully connected layers on pairs of CNN features from non-target classes. We used only 5,000 images but 60,000 pairs to train the kernel. We then tested the kernel on 10,000 image pairs from 1,000 images of unseen target classes, and the results are reported in Table \ref{tab:KernelGeneralization}. The performance was surprisingly high for binary classification, and still encouraging for the 10 class sets. Among the latter, digit comparison was better generalized based on learning to compare alphabet. This was expected because generalization in comparing pen strokes of scripts is inherently easier than features of natural objects.

The kernel took only between 1 to 3 hours to train on a hexa-core 16GB RAM machine with a 2000 CUDA\textregistered core GPU.

\begin{table}[!h]
\caption{Kernel generalization AUC on target (test) classes for training all layers on non-target classes.}
\label{tab:KernelGeneralization}
\centering
\begin{tabular}{| r | r | r | r | r |}
\hline
Images& Pairs& Dogs Cats& MNIST& CIFAR-10 \\ \hline
5,000& 60,000& 0.990& 0.943& 0.868 \\ \hline

\end{tabular}

\end{table}

\subsection{Kernel visualization}

We used visualization of class separation to gain more insights into the relative performance of DESK and its alternative, which is to use CNN features. Tools such as t-SNE are available to visualize class separation for high dimensional data, by projecting the data into two dimensions where their pair-wise distances in small neighborhoods are representative of their distances in the original space~\cite{tSNE}. One can also establish a correspondence between explicit features of a CNN and implicit features of a kernel by using columns of the kernel's gram matrix in lieu of features.

\begin{figure*}
\centering
\includegraphics[width=.9\textwidth, keepaspectratio]{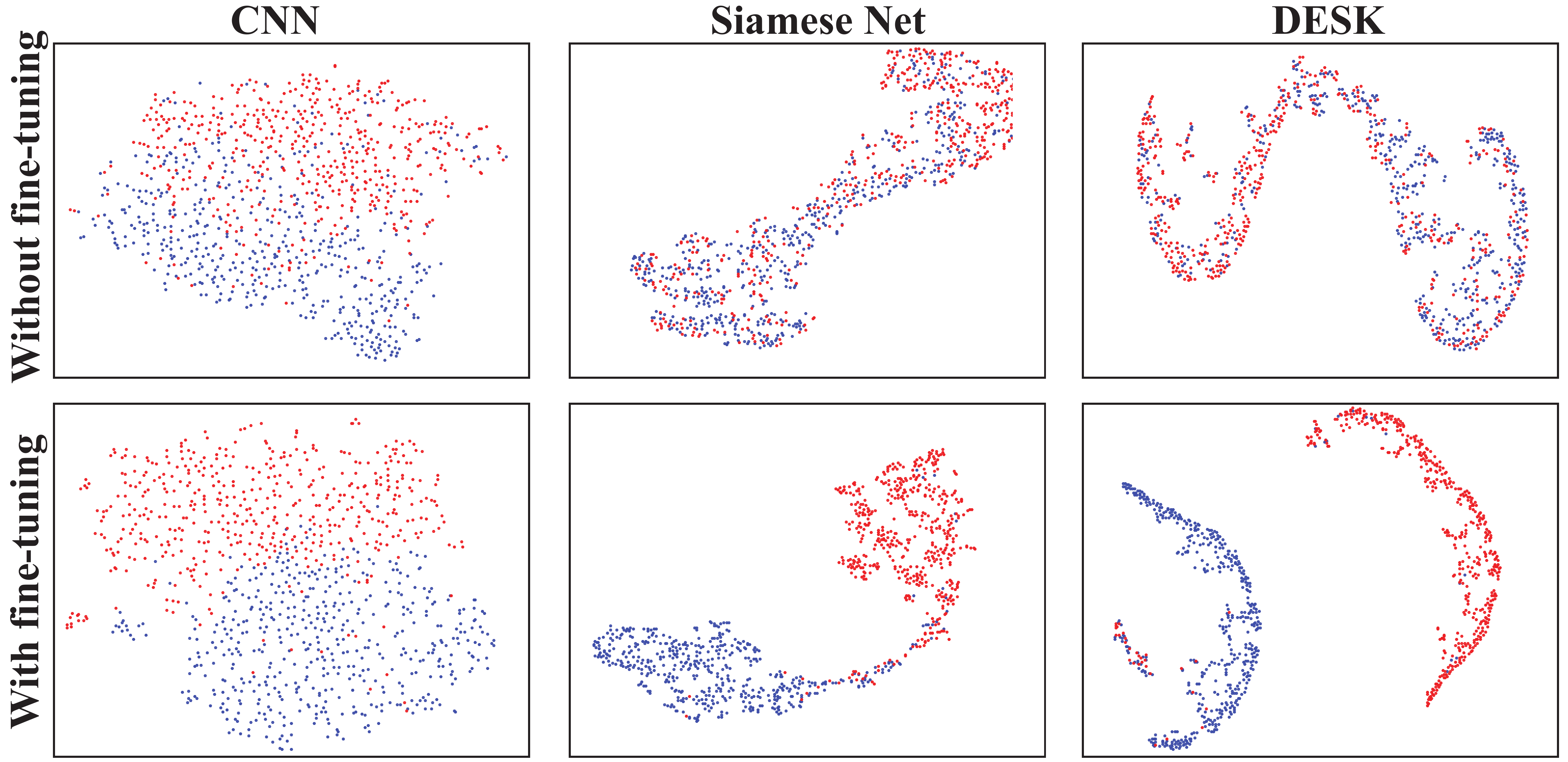}
\caption{Class separation of Dogs vs. Cats for CNN features, Siamese net, and DESK (Siamese net with post-processing) for the two transfer learning schemes (all plots are for 1000 testing samples of target classes).}
\label{fig:tSNE1}
\end{figure*}

\begin{figure*}
\centering
\includegraphics[width=.9\textwidth, keepaspectratio]{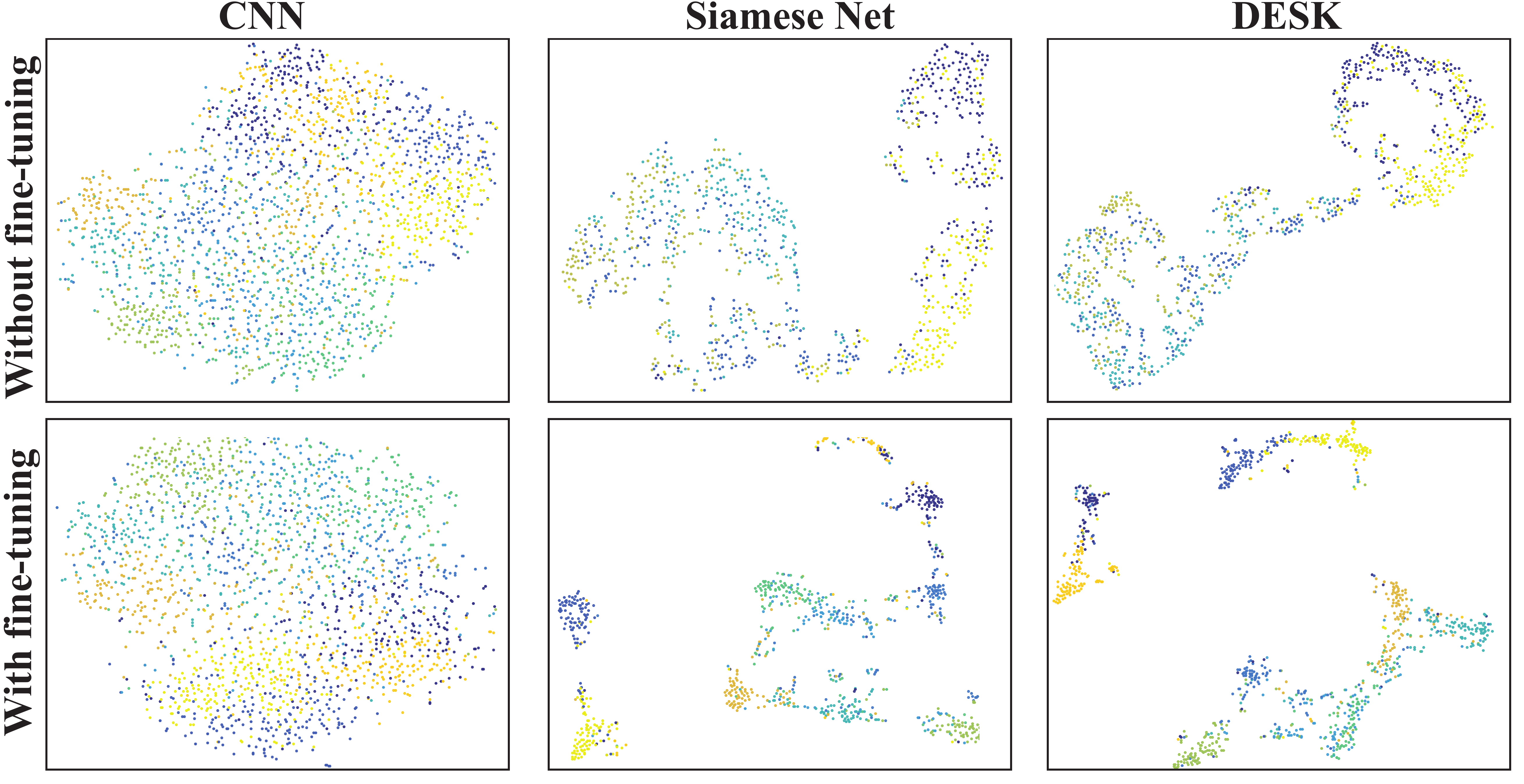}
\caption{Class separation of CIFAR-10 object classes for CNN features, Siamese net, and DESK (Siamese net with post-processing) for the two transfer learning schemes (all plots are for 2000 testing samples of target classes).}
\label{fig:tSNE2}
\end{figure*}

We tried to gain insight into any inherent advantage of DESK over CNN, impact of fine-tuning, and the advantage of using the co-incidence matrix derived from the output of the Siamese net using post-processing. For this, we visualized CNN features, Siamese net output, and DESK output (Siamese with post-processing) for both full-transfer (without fine-tuning) and conv-transfer (with fine-tuning). These results are shown in Figures~\ref{fig:tSNE1} and~\ref{fig:tSNE2}, and are very striking. DESK seems to do a much better job at separating the classes compared to CNN even before fine-tuning. After fine-tuning, the results are remarkably well-separated with only small clusters of confusion.

\subsection{Classification performance using DESK in an SVM}

We now report our main result, which is classification accuracy using a small number of training samples from target classes.

\subsubsection{Conv-transfer}

For the conv-transfer learning scheme where only the convolutional layers were trained on the non-target classes, we got some very encouraging recognition results. After freezing the convolutional layers, we trained the fully connected layers and the SVM on image pairs generated from only 500, 1,000 or 5,000 training images from target classes. This is very frugal considering that most reported work on NIST and CIFAR datasets used around 50,000 training images. For 500 samples, we generated 10,000 image pairs, and for the other two cases we used 60,000 pairs (half of them from the matched classes) for training DESK. Note that the SVM had a slack penalty as a hyper-parameter, which was fixed for testing using a validation subset taken from the training set. We then tested the SVM on 5,000 test images from the target classes. We compared these results with previously reported classification results for the test data sets as well as some obvious alternative transfer learning schemes that use a small number of training examples from target classes. These included use of pre-trained CNN features on non-target classes passed to an SVM via a static kernels (linear and RBF) \cite{huang06}, or coupling pre-trained CNN features with new fully connected layers and fine-tuning on a small number of examples from target classes \cite{alexey15}. DESK, which essentially \emph{learns} a kernel on top of CNN features, generally outperformed these methods in 500 to 5,000 sample range as shown in Table \ref{tab:MasterComparison}. It was observed that the performance of fine-tuned CNN goes up faster with number of training samples than that of DESK, possibly because CNN learns class-appearance-specific weights, while DESK learns comparison-specific weights. Pre-training in our experiments was done using a large number of labeled samples from the mentioned non-target classes.

\begin{table*}
\caption{Classification accuracy vs. number of training samples from target classes for DESK and its alternatives, including the best reported results for MNIST\cite{bachman14} and CIFAR-10\cite{alexey15}.}
\label{tab:MasterComparison}
\centering
\begin{tabular}{|l|c|c|c|c|c|c|c|c|c|}
\hline
Dataset & \multicolumn{3}{c|}{Dogs Cats} & \multicolumn{3}{c|}{MNIST} & \multicolumn{3}{c|}{CIFAR-10} \\ \hline
\# Training samples from target classes & 500 & 1,000 & 5,000 & 500 & 1,000 & 5,000 & 500 & 1,000 & 5,000 \\ \hline
Previously published best results & None&None&None& .976  & .978 &.981 &None&\multicolumn{2}{c|}{.774 at 4,000} \\ \hline
CNN features + linear kernel + SVM & .963 & .965 & .967 & .937 & .946 & .971 & .600 & .628 & .744 \\ \hline
CNN features + RBF kernel + SVM & .963 & .968 & .973 & .926 & .956 & .978 & .661 & .705 & .758 \\ \hline
CNN features + fine-tuning & .810 & .890 & .976 & .797 & .881 & .952 & .617 & .721 & \textbf{.840} \\ \hline
DESK + SVM & \textbf{.996} & \textbf{.995} & \textbf{.997} & \textbf{.989} & \textbf{.995} & \textbf{.993} & \textbf{.758} & \textbf{.774} & .809 \\ \hline
\end{tabular}

\end{table*} 
%\tablefootnote{Results for CIFAR-10 published in \cite{alexey15} were for 4,000 samples, which we interpret to be close to our experiment with 5,000 samples. We achieve the same accuracy as \cite{alexey15} with only 1,000 samples.}

These results are quite encouraging in the sense that with as few as $1,000$ samples, the recognition rates are close to state of the art such as $0.998$ for MNIST~\cite{bestmnist} and $0.84$ for CIFAR-$10$~\cite{Hinton1} using CNNs (with ReLU and dropout) that were trained for classification on approximately $50,000$ samples. These results outperform previously published results for a similar number of samples on these datasets as well as other transfer learning techniques that we implemented.

By comparing Tables \ref{tab:KernelValidation} and \ref{tab:MasterComparison} it is clear that although the kernel accuracy decreased when the number of training images (not pairs) was increased, the classification accuracy either increased or stayed about the same. Thus, although the kernel performed worse when trained on pairs taken from a larger set of images, it perhaps learned a better representation of the data and task, thus aiding the SVM in testing performance.

\subsubsection{Full-transfer}

We next report classification accuracy using a full-transfer kernel in Table \ref{tab:SVMFullTransfer}. That is, even the fully connected layers of the kernel were trained using pairs from non-target classes. Using this transferred kernel, only the SVM was trained on target classes. For dogs vs. cats and digits, the performance is remarkably high considering that only training the SVM on the target classes. This, along with results of Table \ref{tab:KernelGeneralization}, show that a general image similarity can be learned without even fine-tuning the weights of DESK with only using class labels as information.

\begin{table}[!h]
\caption{SVM classification testing accuracy when the entire kernel was transferred from non-target classes.}

\label{tab:SVMFullTransfer}
\centering
\begin{tabular}{| r | r | r | r | r |}
\hline
Images& Pairs& Dogs Cats& MNIST& CIFAR-10 \\ \hline
5,000& 60,000& 0.987& 0.940& 0.607 \\ \hline

\end{tabular}

\end{table}

\section{Discussion and Conclusions}

In this paper, we showed that a Siamese deep neural network architecture that takes two images and estimates their similarity can be used in a trainable kernel -- DESK. We showed that such a network can be trained to capture similarity between two images using supervised learning, where such similarity can be defined by an image pair belonging to the same class, no matter what that class is. With specific modifications to the resultant gram matrix, it can be used in an SVM for classification.

DESK lends itself to transfer learning from non-target classes. With a few hundred to a few thousand training images from target classes, this kernel can be trained to give classification accuracy competitive with traditional CNNs trained on tens of thousands training images to predict target classes. That is, it generalizes to estimating similarity of completely unseen classes, or with only a small amount of training data from the target classes. Consequently, its training time is quite less as well; hours, instead of days. When the entire kernel is pre-trained wholly on tens of thousands of samples from non-target classes, then only the SVM needs to be trained on target classes, which takes only a few minutes. Thus, its main advantage is to be able to train a classifier with far fewer samples and training time on target classes than traditional CNNs in cases when a lot of labeled data is not available.

The main insight from this study is that learning to compare image pairs generalizes better with far fewer samples of target classes than learning to classify. This might be due to offloading the need to code appearance to support vectors instead of storing appearance code templates in the weights of the higher layers in a Siamese architecture. Surprisingly, this can be done using just class labels as information from non-target classes. Using richer semantic information than class labels may lead to even better kernels.

The main disadvantage of DESK is its testing time, as the kernel needs to be computed as a gram matrix of all pairs of testing and training samples, although internally SVM packages only choose to use the columns corresponding to the training support vectors. This load can be reduced if we compute the kernel on pairs of testing samples and only those training samples that correspond to support vectors, which is not allowed in most SVM packages, although it is easy to implement. Moreover, the convolutional part of both support vectors and test samples can be computed only once (separately for each image instead of all pairs of images) due to the disjoint (although Siamese) architecture up to the fully connected layers. Then, only the outputs of fully connected layers need to be computed on all image pairs.

Ways to ensure symmetry of output with respect to the inputs or its positive semi-definiteness right out of the network will be useful to eliminate some of the post-processing that we used. Different variations of SVM and CNN can also be tried to optimize the performance. For example, it may be possible to improve the performance by training the neural network in a deeply supervised framework \cite{lee2014deeply}. The meaningfulness of the kernel can perhaps be further improved for object classification on datasets like CIFAR-10 by not including those pairs from CIFAR-100 that belong to related classes. This is because classes such as dogs and wolves are neither similar nor dissimilar. That is, they are neither the same, nor as far apart as an airplane and a dog. This is where a more granular notion of semantics such as word relational hierarchies may come in handy.

\bibliography{ICML_DESK}
\bibliographystyle{icml2016}

\end{document}